\documentclass[runningheads,a4paper]{llncs}
\usepackage{wrapfig}
\usepackage{amssymb}
\usepackage{graphicx}
\usepackage{hyperref}

\usepackage{url}
\urldef{\mailsa}\path|{alfred.hofmann, ursula.barth, ingrid.haas, frank.holzwarth,|
\urldef{\mailsb}\path|anna.kramer, leonie.kunz, christine.reiss, nicole.sator,|
\urldef{\mailsc}\path|erika.siebert-cole, peter.strasser, lncs}@springer.com|    

\usepackage{afterpage}
\usepackage{amssymb,amsmath,amstext,float}
\usepackage{graphicx}
\usepackage{color}
\usepackage{multicol}
\usepackage{verbatim}
\usepackage{subfig}
\usepackage{verbatim}
\usepackage{authblk}
\usepackage{nameref}

\usepackage{algorithm}
\usepackage{algorithmic}

\usepackage{array} 
\newcolumntype{P}[1]{>{\centering\arraybackslash}p{#1}} 
\newtheorem{step}{Claim}[section]

\newcommand{\mv}[1]{\mathbf{#1}}

\newcommand{\bstep}{\begin{step}}
\newcommand{\estep}{\end{step}}
\newcommand{\blem}{\begin{lemma}}
\newcommand{\elem}{\end{lemma}}
\newcommand{\brem}{\begin{remark}}
\newcommand{\erem}{\end{remark}}
\newcommand{\bthm}{\begin{theorem}}
\newcommand{\ethm}{\end{theorem}}
\newcommand{\beqn}{\begin{equation}}
\newcommand{\eeqn}{\end{equation}}
\newcommand{\eeq}{\end{equation}}
\newcommand{\beq}{\begin{equation}}
\newcommand{\eitem}{\end{itemize}}
\newcommand{\bitem}{\begin{itemize}}
\newcommand{\eenum}{\end{enumerate}}
\newcommand{\benum}{\begin{enumerate}}

\begin{document}

\mainmatter  

\title{Norm-preserving Orthogonal Permutation Linear Unit Activation Functions (OPLU)}

\titlerunning{Orthogonal Permutation Linear Unit Activation Functions (OPLU)}

%
%
\author{Artem Chernodub
\thanks{a.chernodub@gmail.com}
\and Dimitri Nowicki
\thanks{nowicki@nnteam.org.ua}}
\authorrunning{A.N. Chernodub, D.V. Nowicki}

\institute{Institute of MMS of NASU, Center for Cybernetics, 42 Glushkova ave.,  Kiev, Ukraine 03187}

%
%

\toctitle{Lecture Notes in Computer Science}
\tocauthor{Authors' Instructions}
\maketitle

\bibliographystyle{unsrt}

\begin{abstract}
We propose a novel activation function that implements piece-wise orthogonal non-linear mappings based on permutations. It is straightforward to implement, and very computationally efficient, also it has little memory requirements. We tested it on two toy problems for feedforward and recurrent networks, it shows similar performance to tanh and ReLU. OPLU activation function ensures norm preservance of the backpropagated gradients; therefore it is potentially good for the training of deep, extra deep, and recurrent neural networks.
\end{abstract}

\section{Introduction}

Deep neural networks become deeper and deeper. Early DNNs had 4-6 hidden layers \cite{Hinton-2006}, \cite{ciresandeep}. Winner of ILSVRC-2012 (Large Scale Visual Recognition Challenge) AlexNet has 8 layers \cite{krizhevsky2012imagenet}. Winner-2014, VGGNet, has 19 layers \cite{chatfield2014return}, a recent winner, ResNet \cite{he2015deep} has 152 layers. It seems that stacking more non-linear layers produce more freedom for obtaining better performance. The main problem of training deep feedforward networks is vanishing/exploding gradients effect \cite{Bengio-1994}. The same problem exists in recurrent networks, which essentially can be represented as unfolded back through time deep networks with shared weights. One can say that recurrent networks suffer from the vanishing/exploding gradients effect even more because of multiplications on of the same weight matrices during forward and backward passes. Usually, architectural methods are used to prevent vanishing/exploding gradient in RNNs: NARX networks have short links between output error and “old” weights unfolded back in time \cite{Cardot-2011}, LSTM has a special structure of input and forgetting gates which produce a constant error carousel \cite{Hochreiter-1997}, for Echo State networks \cite{Jaeger-2012} only the last feedforward layer is modified during the training and so on. We also mention such an alternative to temporal neural networks as hierarchical sequence processing with auto-associative memories  \cite{kussul1991} that use distributed coding.
Greedy layer-wise pre-training of layers using RBM and autoencoders made a revolution in feedforward deep networks training feedforward deep networks but it still is numerical resources-demanding. “Smart” initialization of DNN’s weights is the current topic of research \cite{glorot2010understanding}, \cite{mishkin2015all}, \cite{le2015simple}, \cite{krahenbuhl2015data}. In \cite{saxe2013exact} orthogonal initial conditions on weights for deep feedforward networks were proposed. It was hypothesized that orthogonality of weight matrices produces a similar effect to unsupervised pre-training that leads to faithful propagation of gradients and faster convergence of training. From the vanishing/exploding gradient perspective this makes sense since the orthogonal matrix preserves the norm of backpropagated gradients. However, traditional activation functions break the orthogonality of backpropagated flow that prevents preserving of gradient norms. Among the vast empirical research on improvement of DNN training methods, search of "good" non-linear activation functions plays an important role, for example, ReLU \cite{glorot2011deep}, Maxout \cite{goodfellow2013maxout}, ELU \cite{clevert2015fast}. In this paper, we propose a novel nonlinear activation function, which is assumed to be used together with the orthogonal weight matrices that may help the training of extremely deep neural networks.

\section{Backpropagaton Mechanics}

Consider a multilayer perceptron (MLP) that has $N$ layers. MLP's $n$-th layer receives postsynaptic activation from the previous layer $\mv{z}^{(n-1)} $ ($\mv{z}^{(0)} $is an input data vector $x$) and produces a new postsynaptic activation $\mv{z}^{(n)} $:

\begin{equation}
  \begin{array}{l}
\mv{a}^{(n)} =\mv{z}^{(n-1)} \mv{w}^{(n)} +\mv{b}^{(n)} ,\\ \mv{z}^{(n)} =f(\mv{a}^{(n)}),
\end{array}
\label{eq1_2}
\end{equation}

where $\mv{w}^{(n)} $is a matrix of weights, $\mv{a}^{(n)} $ is known as a ``pre-synaptic activation'', $f(\cdot )$ is a nonlinear activation function. After processing of all network's layers and producing the output $\mv{y}$, target error $E(\mv{y}(\mv{w}))$ is calculated according to chosen error function $E(\cdot )$. To train the neural network using a gradient-based optimization algorithm we have to calculate derivatives of error function subject to to network's weights $\frac{\partial E}{\partial \mv{w}^{(n)} } $ , $n=1,...,N$ for all network's layers. Standard chain rule-based backpropagation is a common choice for this task. Intermediate variables $\delta ^{(n)} \equiv \frac{\partial E}{\partial \mv{a}^{(n)} } $ are called ``local gradients'' or simply ``deltas''; they are usually introduced for convenience. If deltas are available for specific layer $n$ then corresponding immediate derivatives can be calculated easily: $\frac{\partial E}{\partial \mv{w}^{(n)} } =(\mv{z}^{(n)} )^{T} \delta ^{(n)}$.

For the last layer $\delta ^{(N)} $ is an error residual, for the intermediate layers deltas are incrementally calculated according to very famous backpropagation formula:

\begin{equation}
\delta _{j}^{(n-1)} =f'(a_{j}^{(n-1)} )\sum _{i}\mv{w}_{ij}^{(n)}  \delta _{i}^{(n)}  
\label{eq4}
\end{equation}

Let's write this equation in a matrix form:

\begin{equation}
\delta ^{(n-1)} =\delta ^{(n)} (\mv{w}^{(n)} )^{T} diag(f'(\mv{a}^{(n-1)} )),
\label{eq5}
\end{equation}

where $diag$ converts a vector into to a diagonal matrix. In particular, for  \textit{ReLU} activation function  $f(a_{j} )=\max (a_{j} ,0)$ if we denote $\mv{D}^{(n)} =diag(f'(\mv{a}^{(n-1)} ))$, we get

\begin{equation}
\mv{z}^{(n)} =\mv{a}^{(n)} \mv{D}^{(n)} 
\label{eq6}
\end{equation}

for the forward pass and

\begin{equation}
\delta ^{(n-1)} =\delta ^{(n)} (\mv{w}^{(n)} )^{T} \mv{D}^{(n)} 
\label{eq7}
\end{equation}

for the backward pass where $\mv{D}^{(n)} $ matrix contains either zeros or ones on the diagonal. Equation \eqref{eq5} may be rewritten using the Jacobian matrix $\mv{J}^{(n)} =\frac{\partial \mv{z}^{(n)} }{\partial \mv{z}^{(n-1)} } $:

\begin{equation} 
\delta ^{(n-1)} =\delta ^{(n)} \mv{J}^{(n)} 
\label{eq8}
\end{equation}

where

\begin{equation} 
\mv{J}^{(n)} =(\mv{w}^{(n)} )^{T} diag(f'(\mv{a}^{(n-1)} )).
\label{eq9}
\end{equation}

Now we can see an intuitive understanding of exploding/vanishing gradients problem that was proposed and deeply investigated in classic \cite{Hochreiter-1991}, \cite{Bengio-1994} and modern papers \cite{Pascanu-2012}, \cite{Bengio-2013}. As it follows from \eqref{eq8}, the norm of the backpropagated deltas strongly depends on the norm of the Jacobians. Moreover, they actually are product of Jacobians: $\delta ^{(n-k)} =\delta ^{(n)} \mv{J}^{(n)} \mv{J}^{(n-1)} ...\mv{J}^{(n-k+1)}$.

In practice, Jacobians are more likely to be less than 1 because norms of two factors \eqref{eq9} often are tending to be less than 1. For the first factor, usually $\left\| \mv{w}(n)^{T} \right\| <1$ because large norms leads to non-robust behavior of the neural network. One can easily remember popular weight-decay regularization for neural networks that prevents increasing the weights during the training. 

As for the second factor of \eqref{eq9} $\mv{D}^{(n)} =diag(f'(\mv{a}^{(n-1)} ))$, it's $L_{2}\ $norm is equal to the absolute largest eigenvalue; in standard case if we have a real-valued diagonal matrix it's norm is simply the largest element in $f'(\mv{a}^{(n-1)} )$. The maximum value of derivative for \textit{tanh} is $1$, for \textit{sigmoid} it is ${\raise0.7ex\hbox{$ 1 $}\!\mathord{\left/{\vphantom{1 4}}\right.\kern-\nulldelimiterspace}\!\lower0.7ex\hbox{$ 4 $}} $, so $\left\| \mv{D}^{(n)} \right\| \le 1$ and $\left\| \mv{D}^{(n)} \right\| \le {\raise0.7ex\hbox{$ 1 $}\!\mathord{\left/{\vphantom{1 4}}\right.\kern-\nulldelimiterspace}\!\lower0.7ex\hbox{$ 4 $}} $ respectively. As for \textit{ReLU} function, in the most cases $\left\| \mv{D}^{(n)} \right\| =1$ . Indeed, for ReLU, the largest element in $f'(\mv{a}^{(n-1)} )$ is not 1 if and only if all elements in $f'(\mv{a}^{(n-1)} )$ are zeros. 

At the same time, even if both factors in \eqref{eq9} have norm 1, it still not guarantee norm 1 of $\mv{J}^{(n)} $. For example, if we have $\mv{A}=\left(\begin{array}{cc} {1} & {0} \\ {0} & {0} \end{array}\right)$, $\|\mv{A}\|=1$ and $\mv{B}=\left(\begin{array}{cc} {0} & {0} \\ {1} & {0} \end{array}\right)$, $\|\mv{B}\|=1$ we get $\mv{C}=\mv{AB}=\left(\begin{array}{cc} {0} & {0} \\ {0} & {0} \end{array}\right)$, $\|\mv{C}\|=0$. In practice, after passing the ReLUs norm of gradient $\delta ^{(n)} $ usually becomes smaller. 

The sufficient condition for strict  preservation of the norm of backpropagated gradients \eqref{eq8} is orthogonality of Jacobian matrices \eqref{eq9}. In theoretical work \cite{saxe2013exact} a new class of random orthogonal initial conditions on weights for deep feedforward networks was proposed. It was hypothesized that such orthogonality of weight matrices produces a similar effect to unsupervised pre-training that leads to faithful propagation of gradients and faster convergence of training. At the same time, in the mentioned paper the theoretical analysis and experiments were provided for linear case. In this way, activation functions $f(\cdot )$ are linear functions and therefore the diagonal matrix in Jacobian \eqref{eq9} becomes simply a unity matrix. Thereby, a Jacobian becomes simply a transposed weights matrix, $\mv{J}=\mv{w}_{rec}^{T} $. Orthogonal initialization of weights is an active area of research in Deep Learning community \cite{mishkin2015all}, \cite{le2015simple}, \cite{krahenbuhl2015data}. In \cite{Arjo-2015} a solution based on unitary matrices is proposed.

Meanwhile, using common-known non-linear activation functions breaks the orthogonality of Jacobians even if weight matrices are orthogonal and prevents norm preserving in the backpropagation flow. Activation function that provides orthogonal mapping in a standard neural network's setup where all neurons are independent of each other is not known yet. The obvious solution $\mv{z}=abs(a)$ , unfortunately, is not suitable because it is not a monotonic function and therefore it shows poor convergence properties. In this work, we propose a novel activation function called Orthogonal Permutation Linear Units (OPLU) that ensures the orthogonality of non-linear mapping and acts on neurons in a pair-wise manner.

\section{Orthogonal Permutation Linear Unit activation function (OPLU)}

Activation function produces a vector of postsynaptic values $\mv{z}$ of the same dimensionality for a vector of presynaptic values $\mv{a}$. Suppose we have a neural network's layer with an even number of neurons. Then we may define a list of neuron's pairs; for each pair of input presynaptic values $\left\{a_{i} ,a_{j} \right\}$ we get a pair of neuron's outputs $\left\{z_{i} ,z_{j} \right\}$ according to the following rule: 

\begin{equation}
\left(\begin{array}{c} {z_{i} } \\ {z_{j} } \end{array}\right)=\left(\begin{array}{c} {\max (a_{i} ,a_{j} )} \\ {\min (a_{i} ,a_{j} )} \end{array}\right).
\label{eq12}
\end{equation}

Actually, we perform permutations of pairs of presynaptic values under the certain conditions: $\left(\begin{array}{cc} {z_{i} } & {z_{j} } \end{array}\right)^{T} =\left(\begin{array}{cc} {a_{i} } & {a_{j} } \end{array}\right)^{T} $ if $a_{i} \ge a_{j} $ and $\left(\begin{array}{cc} {z_{i} } & {z_{j} } \end{array}\right)^{T} =\left(\begin{array}{cc} {a_{j} } & {a_{i} } \end{array}\right)^{T} $ else (Fig. 1, left). 

\begin{wrapfigure}{l}{0.5\textwidth}
\begin{center}
\includegraphics[width=0.48\textwidth]{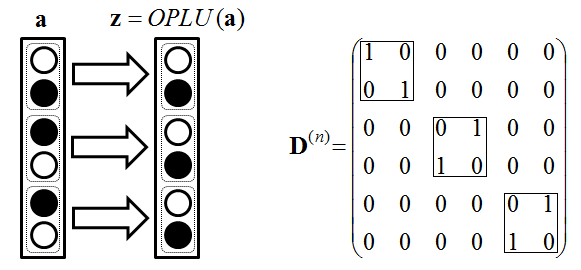}
\end{center}
\caption{Orthogonal Permutation Linear Unit (OPLU) activation function in action(left) and its derivative (right).}
\label{fig1}
\end{wrapfigure}

This 2D mapping has a couple of interesting properties that makes it promising for using as an activation function in neural networks.  First, it is non-linear and continuous. Second, similarly to ReLU, it is a peace-wise linear mapping: forward \eqref{eq6} and backward \eqref{eq7} passes may be expressed as a multiplication of argument on the same matrix $\mv{D}^{(n)} $. Third, this matrix is always orthogonal and, therefore, OPLU activation function is norm-preserving. This is the most important and promising property since now an activation function is not a reason of vanishing or explosion of gradients.

The orthogonality of $\mv{D}^{(n)} $ is easily seen since \eqref{eq12} is equal to the multiplication of a vector of presynaptic values $a$ and one of two orthogonal 2x2 matrices, identity matrix $\left(\begin{array}{cc} {1} & {0} \\ {0} & {1} \end{array}\right)$ or permutation matrix $\left(\begin{array}{cc} {0} & {1} \\ {1} & {0} \end{array}\right)$ and  block-diagonal matrix  whose  blocks are all orthogonal matrices is also an orthogonal matrix.

Finally, it is straightforward to implement, is computationally efficient and has little memory requirements. For implementation using low-level or middle-level language we don't even need to compute the real-valued outputs; what we need is to change integer pointers to the data values in memory.

\begin{tabular}{|p{4.6in}|} \hline 
\textbf{Init}: split all layer's neurons to pairs $\{ p_{k} \} ^{(n)} =\{ (i,j)\} ^{(n)} $, $k=1,...,N_{L} /2$, where $N_{L} $ is a number of neurons in the layer, $n$ is layer's number, $i,j$ are neuron's indexes.\newline \textbf{Forward pass}: for each pair of neurons $p_{k} =(i,j)$, $k=1,...,N_{L} /2$, calculate the next values: \newline $\left[\begin{array}{cc} {z_{i}^{(n)} } & {z_{j}^{(n)} } \end{array}\right]^{T} =\left\{\begin{array}{cc} {\left[\begin{array}{cc} {a_{i}^{(n)} } & {a_{j}^{(n)} } \end{array}\right]^{T} } & {if \ a_{i}^{(n)} \ge a_{j}^{(n)} ,} \\ {\left[\begin{array}{cc} {a_{j}^{(n)} } & {a_{i}^{(n)} } \end{array}\right]^{T} } & {if\ {} a_{i}^{(n)} <a_{j}^{(n)} .} \end{array}\right. $\newline \textbf{Backward pass}: \newline for each pair of neurons $p_{k} =(i,j)$, $k=1,...,N_{L} /2$, calculate the previous deltas:\newline $\left[\begin{array}{cc} {\delta _{i}^{(n)} } & {\delta _{j}^{(n)} } \end{array}\right]^{T} =\left\{\begin{array}{cc} {\left[\begin{array}{cc} {\stackrel{\frown}{\delta }_{i}^{(n+1)} } & {\stackrel{\frown}{\delta }_{j}^{(n+1)} } \end{array}\right]^{T} } & {if \ a_{i}^{(n)} \ge a_{j}^{(n)} ,} \\ {\left[\begin{array}{cc} {\stackrel{\frown}{\delta }_{j}^{(n+1)} } & {\stackrel{\frown}{\delta }_{i}^{(n+1)} } \end{array}\right]^{T} } & {if \ a_{i}^{(n)} <a_{j}^{(n)} .} \end{array}\right. $\newline where\newline $\stackrel{\frown}{\delta }_{m}^{(n+1)} =\sum _{l}\mv{w}_{lm}^{(n+1)}  \delta _{l}^{(n+1)} ,_{} m=1,...,N_{L} .$ \\ \hline 
\end{tabular}
Surely, it is possible to use permutations of orders more than 2. However, it doesn't seem to be useful for practice because high interconnectivity between neurons may lead to overfitting. Actually, the core idea of a popular regularization method ``dropout`` \cite{Srivastava-2014} is to prevent such interconnectivity as much as possible. We suppose that pair-wise interconnectivity between the neurons is a minimal payment for strict orthogonality of mapping's derivative.

\section{Experiments}
\subsection{MNIST Problem}

As a feasibility check, first we tried to train the feedforward network at the MNIST problem. We used the ``LeNet'' convolutional network. It is a standard out-of-box Caffe's example, it's architecture is [conv 5x5]-[pool max]-[conv 5x5]-[pool max]-[full connected]-[ReLU]-[full connected]-[softmax]. Nets were trained using the default parameters: Stochastic Gradient Descent (SGD) algorithm, training speed $\alpha =10^{-2} $, momentum $\mu =0.9$, 10,000 iterations. Initial weights were filled by a standard ``Xavier'' method \cite{glorot2010understanding}. We trained a set of 10 networks for each activation function. As we see from Table 1, for OPLU results are very similar to tanh and ReLU. Surprisingly, we were able to exceed the threshold 99\% without any tuning of training hyper-parameters which were optimized for ReLU function. 

\begin{wraptable}{l}{0.3\textwidth}
\centering
\caption{Classification accuracies at MNIST problem for different activation functions.}
\label{table1}
\begin{tabular}{|l|l|l|}
\hline
     & best     & mean    \\ \hline
TanH & 99.16\%, & 99.07\% \\ \hline
ReLU & 99.17\%  & 99.10\% \\ \hline
OPLU & 99.16\%, & 99.06\% \\ \hline
\end{tabular}
\end{wraptable}

Our Caffe's implementation of OPLU function is available for download here \url{https://github.com/achernodub/oplu\_caffe.git}.

\subsection{Adding problem}   

We trained a recurrent network at the ``Adding'' problem. It is a synthetic problem for testing the ability of the neural network to capture the long-term dependencies in data \cite{Pascanu-2012}, \cite{le2015simple}, \cite{Arjo-2015}. The input consists of a sequence of random numbers, where two random positions (one in the beginning and one in the middle of the sequence) are marked. The model must predict the sum of the two random numbers after the entire sequence was seen. We trained Simple Recurrent Networks (SRN) \cite{Pascanu-2012} with one hidden layer containing 100 units and a linear output layer. 

\begin{wrapfigure}{r}{0.45\textwidth}
\begin{center}
\includegraphics[width=0.42\textwidth]{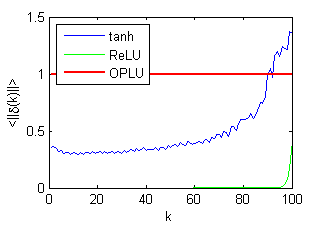}
\end{center}
\caption{ Mean norms of initial backpropagated gradients for SRN, horizon of BPTT $h = 100$. We see that SRN with OPLU activation function and initialized by random orthogonal matrix (red color) has the constant backpropagated gradients.}
\label{fig3}
\end{wrapfigure}

The gradients were obtained using the BPTT method. For tanh and ReLU functions weights were initialized by ``xavier'' method \cite{glorot2010understanding}, for OPLU case the weights were initialized by random orthogonal matrices. To generate them we took a matrix exponential of random skew-symmetric matrices; we casually found out that such initialization works better than the built-in MATLAB's \textit{orth() }function. We trained networks using the SGD, $\alpha =10^{-4} $, $\mu =0.9$, the size of mini-batches is 20. The dataset contains 20,000 samples for training, 1000 samples for validation and 10,000 samples for test. Training process consists 2000 epochs for T=$\{$30,50,70$\}$ and 5000 epochs for T=100, each epoch has 50 iterations. 
\begin{figure}[ht]
\begin{center}
\includegraphics[width=9cm]{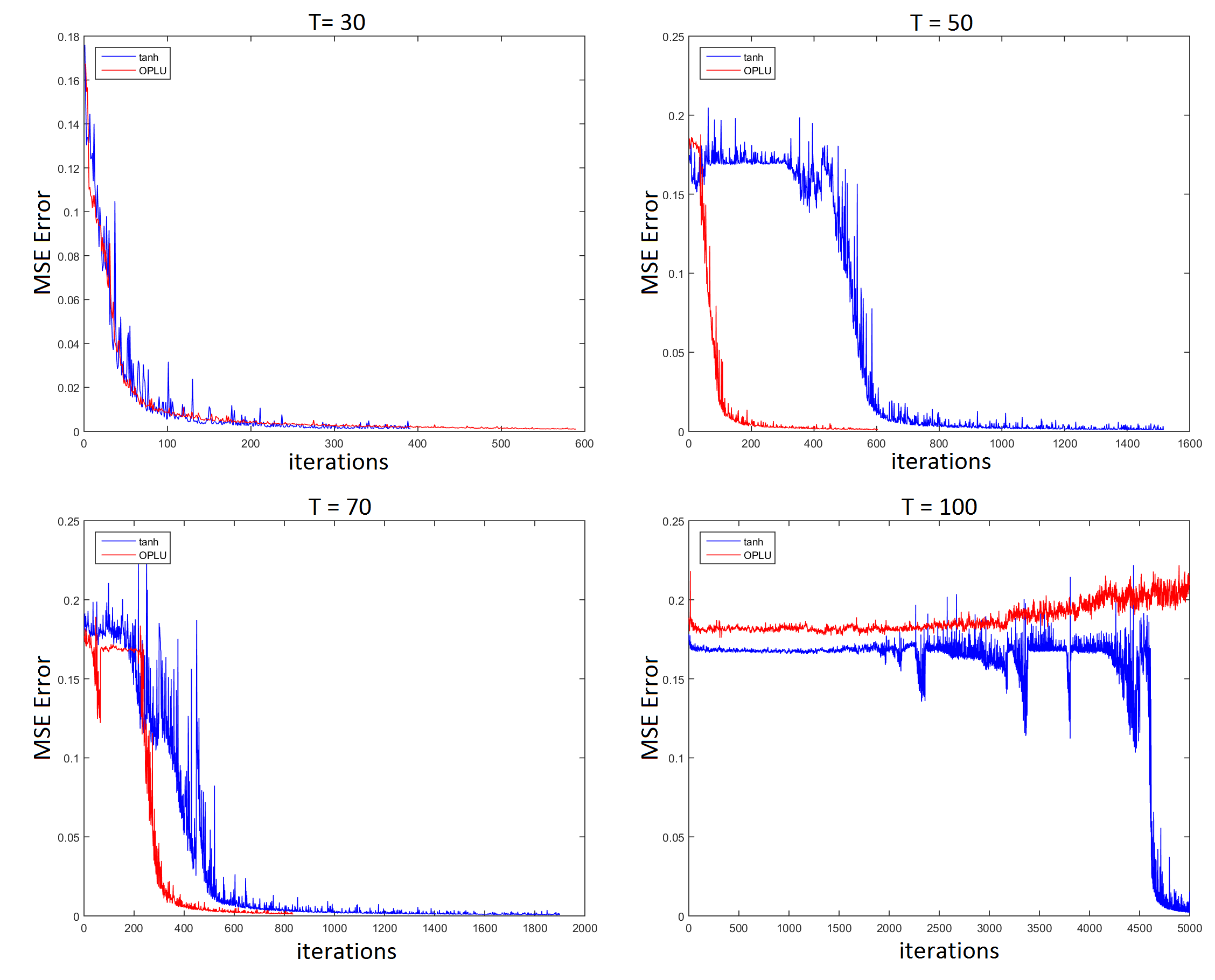}
\end{center}
\caption{Validation MSE error curves for different lengths T during the training for the ``Adding'' problem. Tanh (blue color), OPLU (red color).}
\label{fig4}
\end{figure}

For ReLU activation function we were not able to successfully train the SRN network. It seems that the reason is fast vanishing of gradients for this case (Fig. 2, green). OPLU shows good performance that is similar to tanh for comparatively short sequences T=$\{$30,50,70$\}$. It shows even better performance for the best exemplars from the sets and faster convergence. 

\begin{table}[ht]
\centering
\caption{Rates of  success for different activations for the ``Adding'' problem.}
\label{table2}
\begin{tabular}{|l|ll|ll|ll|ll|}
\hline
     & T= 30                         &         & T=50                         &         & T=70                         &         & T=100                        &         \\ \hline
     & \multicolumn{1}{l|}{best}     & mean    & \multicolumn{1}{l|}{best}    & mean    & \multicolumn{1}{l|}{best}    & mean    & \multicolumn{1}{l|}{best}    & mean    \\ \hline
TanH & \multicolumn{1}{l|}{99.24\%,} & 98.49\% & \multicolumn{1}{l|}{98.90\%} & 98.48\% & \multicolumn{1}{l|}{99.31\%} & 90.46\% & \multicolumn{1}{l|}{98.44\%} & 52.91\% \\ \hline
OPLU & \multicolumn{1}{l|}{99.34\%,} & 98.83\% & \multicolumn{1}{l|}{99.21\%} & 98.70\% & \multicolumn{1}{l|}{99.43\%} & 81.58\% & \multicolumn{1}{l|}{16.33\%} & 15.36\% \\ \hline
\end{tabular}
\end{table}

However, for the T=100 we were not able to successfully train the SRN with OPLU, currently, we can't accurately explain why. The MATLAB code for this experiment is available here: \url{https://github.com/achernodub/oplu\_adding.git}.

\section{Conclusion}

In this study we introduced a new type of piecewise-linear activation function. This function (OPLU) acts pairwise on postsynaptic potentials of network’s layer, and its derivative is an orthogonal operator at every point. This approach is promising thanks to strong and clear mathematical justification that guarantees strict norm preservation for unlimited number of layers if their weight matrices are orthogonal. It is also interesting that an isomorphism could be established between OPLU activation fuction and Maxout \cite{goodfellow2013maxout}.  At current stage of the research we proved feasibility of OPLU activation for small problems for feed-forward convolutional and simple recurrent networks. 

Exploring of its potential and limitations for real-life problems is a subject of our future research.

\section*{Acknowledgments}

We thank FlyElephant (http://flyelephant.net) and Dmitry Spodarets for computational resources kindly given for our experiments.

\bibliography{oplu-bib1}
\end{document}